\documentclass[conference]{IEEEtran}
\IEEEoverridecommandlockouts
\usepackage{cite}
\usepackage{amsmath,amssymb,amsfonts}
\usepackage{algorithmic}
\usepackage{graphicx}
\usepackage{textcomp}
\usepackage{xcolor}
\usepackage{graphicx}
\usepackage{caption}
\usepackage{subcaption}
\usepackage[utf8]{inputenc}
\usepackage{amsfonts}
\usepackage{pgfplots}
\usepackage{makecell}
\DeclareUnicodeCharacter{2212}{−}
\usepgfplotslibrary{groupplots,dateplot}
\usetikzlibrary{patterns,shapes.arrows}
\pgfplotsset{compat=newest}
\usepackage{hhline}
\usepackage[inline,shortlabels]{enumitem}
\usepackage{svg}
\usepackage{adjustbox}
\usepackage{import}
\usepackage{hyperref}
\usepackage{hyperxmp}
\usepackage{amsmath}
\usepackage{comment}
\usepackage{siunitx}
\usepackage{booktabs}
\usepackage{multirow}
\usepackage{threeparttable}
\usepackage{adjustbox}
\usepackage[utf8]{inputenc}
\usepackage{textgreek}
\def\BibTeX{{\rm B\kern-.05em{\sc i\kern-.025em b}\kern-.08em
    T\kern-.1667em\lower.7ex\hbox{E}\kern-.125emX}}

\newcommand{\tp}[1]{\textcolor{red}{{\bf[$^{Themis: }$ ??? #1 ???]}}}

\begin{document}

\title{LoaDiff: Conditional Generation of Electricity Consumption Time Series for Energy Analytics}


\author{
\IEEEauthorblockN{Mariia Baranova\IEEEauthorrefmark{1}\IEEEauthorrefmark{2}\IEEEauthorrefmark{3}, Adrien Petralia\IEEEauthorrefmark{1}\IEEEauthorrefmark{3}, Etienne Le Naour\IEEEauthorrefmark{1},\\ Nathan Etourneau\IEEEauthorrefmark{1}, Guillaume Hofmann\IEEEauthorrefmark{1}, Themis Palpanas\IEEEauthorrefmark{2}}
\IEEEauthorblockA{\IEEEauthorrefmark{1}\textit{EDF R\&D}, Palaiseau, France;
\IEEEauthorrefmark{2}\textit{Universit\'{e} Paris Cit\'{e}}, F-75006 Paris, France}
\IEEEauthorblockA{\IEEEauthorrefmark{3}These authors contributed equally to this work.}
}

\maketitle

\begin{abstract}
The energy transition is reshaping residential electricity consumption through the increasing adoption of distributed generation, electrified appliances, and demand-response programs. 
Understanding these evolving behaviors requires access to granular smart-meter data for applications such as load forecasting, appliance detection, and demand-side flexibility analysis. 
However, such data are subject to strict access restrictions and data-protection regulations. 
Thus, realistic synthetic alternatives are necessary.
In this paper, we introduce \textsc{LoaDiff}, a diffusion-based generative model for year-long, sub-hourly smart-meter load curves. \textsc{LoaDiff} supports flexible conditioning on static household attributes, such as appliance ownership, and dynamic contextual variables, including calendar information and outdoor temperature. We evaluate the model against multiple generative baselines on three residential electricity-consumption datasets. 
Our experiments assess four complementary dimensions: fidelity and diversity, training-record memorization risk, downstream utility for load forecasting and appliance detection, and conditional controllability under alternative temperature conditions. 
The results show that \textsc{LoaDiff} generates realistic and diverse load profiles, achieves a favorable trade-off between generation quality and limited evidence of memorization, preserves information useful for downstream energy applications, and responds coherently to changes in conditioning variables.
\end{abstract}


\section{Introduction}
\label{sec:intro}

Smart meters are now widely deployed worldwide, recording household electricity consumption at regular intervals and generating large-scale datasets for energy analytics~\cite{smart_meter_deployments_ue}. 
Typically collected every 10--30 minutes, depending on the country~\cite{eEnergy_ApplDetection}, these measurements provide detailed insights into residential electricity use. 
Utilities rely on such data for demand forecasting, billing, and the development of services such as personalized feedback tools, time-of-use tariffs, and demand-response programs \cite{eEnergy_ApplDetection,VLDB_TransApp}.  
At this granularity, smart-meter data also support a wide range of data-driven applications, including customer segmentation, non-intrusive load monitoring, and the evaluation of household energy-efficiency actions~\cite{themis_reviewnilm2024,camal_icde,devicescope_icde,nilmformer2025}. 
Thus, smart meters have become a cornerstone of modern energy analytics and power-system planning.

However, individual smart-meter load curves are personal data under GDPR~\cite{gdpr2016}, potentially revealing occupancy schedules, work habits, holidays, or appliance use~\cite{mckenna2012smart}. As a result, access is subject to strict legal and technical safeguards, and advanced analytics often remain confined within utility infrastructures, limiting access for external researchers, regulators, and third-party innovators~\cite{EnergyManagement}.

Synthetic data generation has therefore emerged as a promising alternative: artificial load curves that reproduce the statistical properties of real electricity consumption while not corresponding to any real household.
Early approaches relied on parametric or rule-based simulators combining appliance models with occupancy patterns~\cite{RICHARDSON20081560}. 

\begin{figure}[tb]
\centering
    \includegraphics[width=1\linewidth]{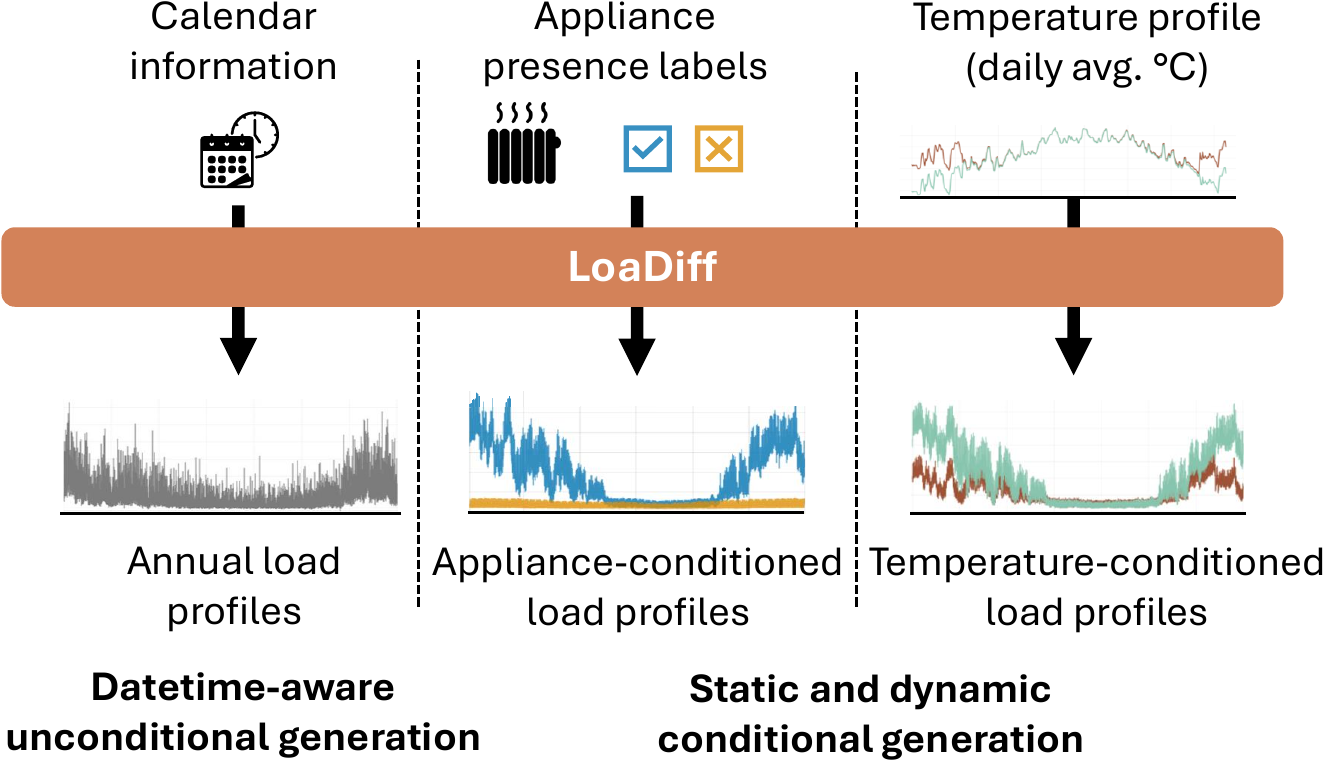}
    \caption{Illustration of \textsc{LoaDiff} generation capabilities: unconditional generation, conditioning on static household attributes, and conditioning on dynamic variables.}
    \label{fig:LoaDiff_intro}
    \vspace*{-0.5cm}
\end{figure} 

More recently, deep generative models, e.g., variational autoencoders~\cite{pan2019data}, generative adversarial networks~\cite{liang2022synthesis} and diffusion models~\cite{energydiff2025}, have been used to learn realistic consumption patterns directly from data. 
However, existing approaches suffer from several limitations: they often focus on short horizons (e.g., day or week) rather than long-term behavior~\cite{liang2022synthesis,thorve2023high,yuan2023synthetic, energydiff2025}; provide limited control over exogenous factors such as weather, calendar effects, or appliance ownership; and rarely offer systematic evaluation of fidelity, downstream utility, and privacy risk.

In this paper, we introduce \textsc{LoaDiff}, a diffusion-based generative model designed for long-horizon, yet fine-grained, smart-meter data.
As illustrated in Figure~\ref{fig:LoaDiff_intro}, \textsc{LoaDiff} generates year-long, sub-hourly load traces while supporting flexible conditioning on both static household attributes, such as appliance ownership, and dynamic factors, such as calendar information and outdoor temperature.
This formulation is intended to support population-level scenario generation while preserving the temporal structure required by downstream energy applications.

We evaluate \textsc{LoaDiff} on two real-world smart-meter datasets and one simulator-backed dataset, against multiple generative baselines.
Our evaluation considers four complementary dimensions:
(i) \emph{fidelity and diversity}, measuring how closely synthetic data reproduce the statistical structure of the target population;
(ii) \emph{privacy risk}, assessing whether generated samples exhibit evidence of training-record memorization via distance-based proximity to the training set;
(iii) \emph{downstream utility}, measuring whether synthetic curves can replace or complement real training data for load forecasting and appliance detection; and
(iv) \emph{conditional controllability}, testing whether generated profiles respond coherently to changes in winter temperatures.
Finally, we release the implementation of \textsc{LoaDiff}~\cite{loadiffcode} together with a large synthetic smart-meter dataset~\cite{loadiffcerdataset} to foster reproducibility and broader experimentation with privacy-aware energy data.

Our contributions are summarized as follows:
\begin{itemize}[label=\textbullet]
\item We introduce \textsc{LoaDiff}, a diffusion-based generative model for long (i.e., year-long) and fine-grained (i.e., sub-hourly) smart-meter load curves. 
Moreover, the model supports conditional generation from both static household covariates and dynamic contextual variables.

\item We conduct an extensive empirical evaluation against multiple generative baselines. Across three datasets, \textsc{LoaDiff} achieves strong fidelity--diversity performance while exhibiting a favorable empirical trade-off between generation quality and limited evidence of training-sample memorization.

\item We assess downstream utility on two practical tasks: load forecasting and appliance detection. 
The results show that \textsc{LoaDiff} samples preserve useful temporal and appliance-specific information, yielding consistently strong performance used as a substitute for real training data, providing effective augmentation.

\item We evaluate conditional controllability through counterfactual temperature scenarios. The generated load profiles tend to respond consistently with expected patterns under winter-temperature shifts, suggesting the model captures a plausible relationship between colder weather and electric-heating demand.

\item We release an open-source implementation of \textsc{LoaDiff} together with a synthetic smart-meter dataset to support reproducible research on energy-consumption modeling and synthetic-data evaluation.

\end{itemize}

\section{Related Work}
\label{sec:relatedwork}

This section reviews related work along three axes relevant to \textsc{LoaDiff}: 
(i) generative models for time series, 
(ii) conditional and controllable generation, and 
(iii) applications of generative models to electricity load curve generation.

\noindent{\bf Generative Models for Time Series.}
Generating realistic time series has been an active research topic for many years. Classical approaches such as ARIMA, Gaussian Mixture models, or hidden Markov models capture seasonal patterns but struggle with complex nonlinear dynamics and heterogeneous behaviors \cite{gershenfeld2018future}. Deep generative models now offer stronger alternatives: VAEs~\cite{timevae2022}, GANs~\cite{timegan2021}, and diffusion-based models~\cite{diffusionts2024} efficiently capture high-dimensional temporal dependencies, making them promising candidates for realistic time-series generation.

\noindent{\bf Conditional Generation and Controllability.}
These methods have enabled synthetic time-series generation in domains where privacy constraints limit data sharing, including finance \cite{wiese2020quant}, mobility \cite{chatterjee2023generating}, and healthcare \cite{yang2023ts}. However, unconditional generation is often insufficient: users typically need control over specific attributes. Conditional generative models address this by incorporating auxiliary information—static covariates \cite{timevqvae2023} or dynamic exogenous variables \cite{timeweaver2024}—enabling controllable generation and scenario analysis. This is especially relevant in the energy domain, where consumption strongly depends on external covariates such as weather, calendar effects, and appliance ownership \cite{grandjean2012review}.

\noindent{\bf Synthetic Load Curve Generation.}
Several works have explored generating synthetic electricity consumption data using deep generative models~\cite{liang2022synthesis,yuan2023synthetic,thorve2023high, energydiff2025}, but typically focus on short time windows (e.g., daily shapes) with limited control over exogenous drivers. \cite{nabil2025syntheticdatasetfrenchelectric} conditions load curve synthesis on outdoor temperature, yet flexible conditioning on multiple household attributes—such as appliance ownership, essential for evaluating energy-efficiency actions or demand-response scenarios—remains lacking. \textsc{LoaDiff} addresses this gap, generating realistic yearly sub-hourly load curves conditioned on both static household characteristics and dynamic variables such as weather and calendar effects.

\begin{figure*}[tb]
    \centering
    \includegraphics[width=0.9\linewidth]{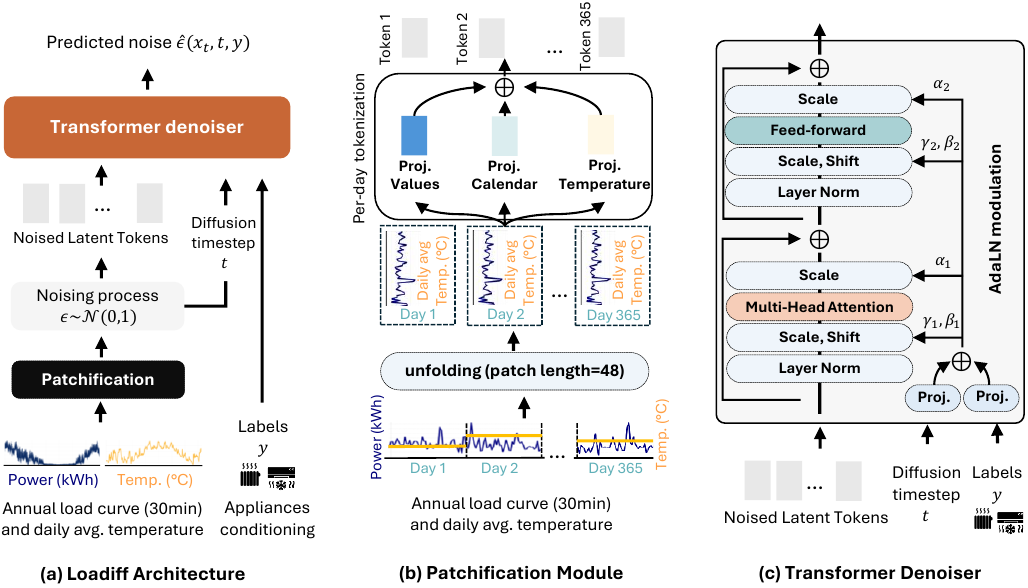}
    \caption{
    Overview of the LoaDiff framework: (a) overall architecture (training phase), (b) detailed view of the patchification module, and (c) detailed view of the Transformer-based denoiser.
    }
    \label{fig:LoaDiffbackbone}
    \vspace{-0.2cm}
\end{figure*}

\section{Problem Formulation}
\label{sec:problem_overview}

This section \begin{enumerate*}[(i)] \item introduces  the notation and formalizes the conditional generation problem addressed by \textsc{LoaDiff}, and then \item  describes the evaluation framework used to assess the quality of the generated synthetic data. \end{enumerate*}

\noindent{\bf Notations and Generation Formulation.}
Consider a univariate time series sample $\mathbf{x} = (x_1, \dots, x_T) \in \mathbb{R}^T$ representing the electricity consumption load curve of a given household. 
Each series is observed over a long horizon of $T$ time steps at a sub-hourly resolution (e.g., $T = 17{,}520$ for one year at 30-minute intervals). 
Each element $x_t$ corresponds to the measured power consumption at time $t$, typically expressed in Watt-hours. Each load curve $\mathbf{x}$ is associated with a set of exogenous variables capturing both static and dynamic information.  We denote by 

\begin{itemize}[label=\textbullet, leftmargin=*, labelindent=0pt]
    \item $\mathbf{s} \in \mathbb{R}^{d_s}$ a vector of static household descriptors (e.g., dwelling type, climate zone, appliance ownership)
    \item $\mathbf{z} = (z_1,\dots,z_T)$ a sequence of dynamic covariates with $z_t \in \mathbb{R}^{d_z}$ (e.g., time-of-day indicators, calendar effects, outdoor temperature, or price signals). 
    \item $\mathbf{c} = (\mathbf{s}, \mathbf{z})$ for the full conditioning information.
    \item $\mathcal{D} = \{(\mathbf{x}^{(i)}, \mathbf{s}^{(i)}, \mathbf{z}^{(i)})\}_{i=1}^{N}$ the training dataset.
\end{itemize}

The objective of \textsc{LoaDiff} is to learn a parametric conditional model $p_\theta(\mathbf{x}\mid \mathbf{s}, \mathbf{z})$ which approximates the true conditional distribution $p(\mathbf{x}\mid \mathbf{s}, \mathbf{z})$. 
Once trained, the model can generate synthetic load curves $\tilde{\mathbf{x}} \sim p_\theta(\mathbf{x}\mid \mathbf{s}, \mathbf{z})$ that are consistent with the conditioning variables.

\noindent{\bf Evaluation of Synthetic Smart-Meter Data.}
Assessing the quality of synthetic load curves $\tilde{\mathbf{x}}$ generated from $p_\theta(\mathbf{x}\mid\mathbf{s},\mathbf{z})$ raises specific methodological challenges. 
Following the synthetic data literature, we evaluate the generated data along three complementary axes, explained below.

    \noindent{(i)} \emph{Fidelity} measures how closely the synthetic samples $\tilde{\mathbf{x}}$ reproduce the statistical properties of real load curves $\mathbf{x}$, including marginal distributions, temporal dependencies, and joint relationships with the conditioning variables $(\mathbf{s},\mathbf{z})$. 
    
    \noindent{(ii)} \emph{Utility} evaluates whether synthetic data can effectively replace real data in downstream tasks. This is assessed by training  models on synthetic samples and measuring their performance on real-world test data (Train on Synthetic, Test on Real framework).
    
    \noindent{(iii)} \emph{Privacy} Privacy examines the extent to which generated samples show evidence of memorizing records from the training dataset $D$. This is particularly important in the context of smart-meter data, where detailed load curves may reveal sensitive information about household behavior. Note that this evaluation does not provide formal privacy guarantees (e.g., differential privacy).

Many smart-meter synthesis works focus on fidelity metrics (e.g., marginal statistics or autocorrelation), with limited utility analysis and cursory privacy checks. 
We evaluate all three dimensions to assess practical utility and empirical privacy risk.

\section{The \textsc{LoaDiff} approach}
\label{sec:proposedapproach}

This section introduces \textsc{LoaDiff}, a conditional generative model for smart-meter load curves.
\textsc{LoaDiff} learns a distribution over individual household electricity trajectories that captures multi-scale temporal patterns, while allowing conditioning on exogenous factors such as calendar features, outdoor temperature, and appliance ownership. 
The following subsections describe the main components of our approach:

\noindent\textbf{Diffusion-based generation.} \textsc{LoaDiff} relies on a denoising diffusion framework, in which the model is trained to progressively remove noise from corrupted time series. At inference time, realistic load curves are generated by iteratively denoising a random noise sequence.
    
\noindent\textbf{DiT-based backbone.} The denoising network is implemented using a Transformer architecture inspired by Diffusion Transformers (DiT)~\cite{peebles2023dit}, adapted to long one-dimensional time series. To preserve daily structure and reduce sequence length, each yearly load curve is reshaped into a two-dimensional representation and partitioned into non-overlapping daily patches.
    
\noindent\textbf{Conditioning on Static and Continuous Covariates.}  This representation maintains temporal correlations while enabling efficient conditioning: dynamic covariates are incorporated at the patch level, while static household attributes are injected through Adaptive Layer Normalization (AdaLN).

The following subsections provide a more detailed description of these components.

\subsection{Diffusion-based generative modeling of load curves}
\textsc{LoaDiff} employs a conditional denoising diffusion probabilistic model (DDPM)~\cite{ho2020denoising}. Following the standard DDPM training procedure, we train a denoiser network $\boldsymbol{\epsilon}_\theta$ to predict the noise added to a clean load curve $\mathbf{x}_0$ at a random timestep $k$. The model is optimized using a simple mean-squared error objective between the predicted and the true noise.

To generate a new sample, we start from pure Gaussian noise $\mathbf{x}_K$ and iteratively apply the learned reverse process, conditioned on variables $\mathbf{c}$, to denoise the sample over $K$ steps. The architecture of the denoiser $\boldsymbol{\epsilon}_\theta$ is described next.

\subsection{A DiT-like Transformer Backbone for Time Series}
\label{subsec:dit_backbone}

The denoising network used in \textsc{LoaDiff} relies on a DiT-like Transformer backbone~\cite{peebles2023dit} operating on a tokenized representation of the load curve and its covariates. 
Directly processing year-long sub-hourly load curves as raw sequences would result in prohibitively long input sequences.  
To maintain computational efficiency while preserving the inherent daily structure of electricity consumption, we reshape each time series from its sequential representation $\mathbf{X} \in \mathbb{R}^{T}$, with $T = 17{,}520 = 365 \times 48$, into an image-like tensor $\mathbf{X} \in \mathbb{R}^{H \times W}$, where $H$ corresponds to the number of days in the year ($365$) and $W$ to the number of half-hour intervals per day ($48$).  
This representation makes the temporal structure more explicit and enables patch-based tokenization similar to Vision Transformers. 
A visualization of the denoiser is shown in Figure~\ref{fig:LoaDiffbackbone}.

\noindent{\bf Patching and Tokenization.} 
The first layer of the architecture performs a \textit{patchification} step that converts the input time series into a sequence of non-overlapping patches. 
Given the reshaped input $\mathbf{X} \in \mathbb{R}^{H \times W}$, this operation produces 
$L = \frac{H}{p_H} \times \frac{W}{p_W}$ patches 
$(\mathbf{x}^{(1)}, \dots, \mathbf{x}^{(L)})$, where 
$\mathbf{x}^{(\ell)} \in \mathbb{R}^{P}$ denotes the $\ell$-th patch flattened into a vector of dimension 
$P = p_H p_W$. In this setting, the input grid is non-square with shape $\mathbf{X} \in \mathbb{R}^{H \times W}$, where $H$ corresponds to the number of days and $W$ to the number of half-hour intervals per day. 
To preserve the daily structure of electricity consumption, non-square patches of size $(p_H, p_W) = (1,48)$ are used. 
Each patch therefore corresponds to a single day of observations, resulting in a fixed number of tokens $L = 365$. 
For each patch, the final token representation is obtained by adding the patchified consumption values, the aligned dynamic covariates, and their interaction features:
$
\mathbf{u}^{(\ell)}
=
\mathbf{x}^{(\ell)}
+
\mathbf{z}^{(\ell)}
+
\phi!\left(\mathbf{x}^{(\ell)},\mathbf{z}^{(\ell)}\right)
\in \mathbb{R}^{d_{\text{patch}}}.
$
These combined patch features are then projected into the Transformer embedding space:
$
\mathbf{h}^{(\ell)}*0
=
\mathbf{W}*{\text{patch}}\mathbf{u}^{(\ell)}
+
\mathbf{b}*{\text{patch}}
\in \mathbb{R}^{d*{\text{model}}}.
$

\noindent{\bf Timestep and Positional Conditioning.}
To inform the network about the stage of the diffusion process, the diffusion timestep $k$ is encoded using a learned embedding $\mathbf{e}_k \in \mathbb{R}^{d_{\text{model}}}$.
This embedding is injected into each Transformer block through Adaptive Layer Normalization (AdaLN). In the AdaLN mechanism, the scale and shift parameters of the normalization layers are modulated by the conditioning vector.

In addition, the denoiser uses a temporal positional encoding $\mathbf{p}^{(\ell)}$ to represent the absolute location of each patch within the yearly horizon. This encoding is constructed solely from the discrete timestamp values associated with the input subsequences. 
More precisely, given a timestamp $t_i$, we extract four calendar features: weekday $t_i^{w}$, day of month $t_i^{d}$, day of year $t_i^{y}$, and month $t_i^{M}$. Each of these variables is mapped onto a periodic representation using its corresponding angular frequency:
$
\theta_i^j = \frac{2\pi t_i^j}{p^j},
\qquad j \in \{w,d,y,M\},
$
where $\{p^w=7,\; p^d=31,\; p^y=365,\; p^M=12\}$ denote the periods used for weekday, day of month, day of year, and month, respectively. 
The resulting four-dimensional calendar representation is then projected into the model space through a $1$D convolution with kernel size $1$.

\noindent{\bf Transformer blocks.}
As illustrated in Figure~\ref{fig:LoaDiffbackbone}(c), the denoiser backbone consists of $B$ stacked Transformer blocks. 
Each block comprises multi-head self-attention and a feed-forward network, both wrapped with residual connections and AdaLN conditioned on the diffusion timestep and static attributes:
\begin{align}
    \mathbf{h}' &= \text{MSA}\bigl(\text{AdaLN}(\mathbf{h}; \mathbf{e}_k, \mathbf{e}_s)\bigr) + \mathbf{h}, \\
    \mathbf{h}^+ &= \text{FFN}\bigl(\text{AdaLN}(\mathbf{h}'; \mathbf{e}_k, \mathbf{e}_s)\bigr) + \mathbf{h}',
\end{align}
where $\mathbf{h} \in \mathbb{R}^{L \times d_{\text{model}}}$ is the token matrix, 
$\mathbf{e}_s$ is an embedding of static household descriptors $\mathbf{s}$, and MSA/FFN denote multi-head self-attention and a position-wise feed-forward network, respectively.
This design allows the model to adapt its denoising behavior based on both the diffusion stage and household characteristics.

\noindent{\bf Output head.}
After the final block, a linear projection maps token representations back to the patch space, yielding an estimate of the noise for each patch: $\widehat{\boldsymbol{\epsilon}}^{(\ell)} = \mathbf{W}_{\text{out}} \mathbf{h}^{(\ell)}_B + \mathbf{b}_{\text{out}}.$
Reassembling the patches gives the full noise prediction $\boldsymbol{\epsilon}_\theta(\mathbf{x}_k, k, \mathbf{c}) \in \mathbb{R}^{T}$.

\subsection{Conditioning on exogenous factors}

\textsc{LoaDiff} supports 
conditioning on exogenous factors, for scenario-based generation and counterfactual analyses.

\noindent{\bf Static Attributes.}
Static descriptors $\mathbf{s}$ (e.g., presence label of an electric heater or an electric vehicle) are embedded into a latent vector $\mathbf{e}_s = f_s(\mathbf{s})$ using a small MLP or embedding lookup for categorical variables.
As in the original DiT paper~\cite{peebles2023dit}, the resulting embedding is injected into every Transformer block via AdaLN, allowing the network to adapt its denoising dynamics to the household profile.

\noindent{\bf Dynamic Covariates.}
Continuous (i.e., time-varying covariates $\mathbf{z}$, such as temperature) and calendar profiles, are incorporated in two complementary ways.
First, as described above, they are concatenated with the consumption values when forming patch features $\mathbf{u}^{(\ell)}$, providing a local, time-aligned view of exogenous drivers.

\noindent{\bf Classifier-free Guidance for Controllability.}
To further improve controllability and robustness, we adopt a classifier-free guidance scheme. 
During training, a fraction $p_{\text{drop}}$ of conditioning information is randomly dropped (replaced by a learned \textsc{null} embedding), and the model learns both conditional and unconditional denoising. 
At sampling time, we can trade off fidelity to the conditioning versus diversity by interpolating between conditional and unconditional predictions, as is standard in diffusion models. 
In practice, this mechanism allows users to enforce strong adherence to specific conditioning scenarios (e.g., a particular heating technology and weather year) while avoiding overfitting to spurious correlations in the training data.

Overall, this DiT-like architecture enables \textsc{LoaDiff} to jointly model long-range temporal dependencies, integrate rich exogenous information, and generate realistic, controllable synthetic load curves over year-long horizons. 




\section{Experimental Evaluation}
\label{sec:expsetup}

\subsection{Datasets}
\label{sec:datasets} 

We evaluate the models on three residential electricity-consumption datasets. 
The first is sourced from the Irish Social Science Data Archive (ISSDA)~\cite{ISSDA}, while the other two are provided by Électricité de France (EDF), the main electricity provider in France.

\subsubsection{CER Dataset}
\label{dataset-desc}

The Irish Commission for Energy Regulation monitored electricity consumption in more than 5,000 homes and businesses between 2009 and 2011 to assess smart-meter performance and its impact on consumer behavior~\cite{CER_2012}.
Our experiments use the residential subset: 4,225 households recorded at 30-minute intervals from July 15, 2009, to January 1, 2011, yielding 4,225 load series of length 25,728.
Participant questionnaires provide household-composition and appliance-ownership information.

\subsubsection{EDF Datasets}

\begin{itemize}[label=\textbullet, leftmargin=*, labelindent=0pt]
    \item \textbf{EDF 1.} This dataset is based on a survey conducted by EDF to better understand its customers and their electricity consumption behavior.  The total power consumption of 2083 houses was recorded every 30min. The dataset contains one year recorded consumption series collected between January 2024 and December 2025. 
    Like the CER dataset, customers filled out a questionnaire with information about household composition, including the appliances in the house.
    \item \textbf{EDF 2 (SMACH).} The second dataset provided by EDF is a synthetic dataset generated with the SMACH agent-based simulation framework, which models human activities and their impact on household electricity consumption. 
It consists of 20K synthetic residential customers, each associated with an electricity consumption curve generated from a detailed household description. 
Each curve is conditioned on several contextual and household-level variables, including household type, household composition, appliance ownership, weather conditions, and other descriptive information available to the simulator. 
\end{itemize}

\subsection{Evaluation Pipeline}
\label{sec:evaluation_pipeline} 

We evaluate \textsc{LoaDiff} against established time-series generative models under the conditioning regimes supported by each method. 
The pipeline 
assesses not only whether the generated load curves resemble real consumption traces, but also whether they preserve task-relevant information and respond meaningfully to controlled changes in the conditioning variables.

\subsubsection{Baselines}\label{sec:baselines}
The comparison includes generative models spanning three regimes:
\begin{enumerate*}[(i)]
\item \emph{unconditional} generation,
\item \emph{static-conditioned} generation, based on time-invariant household attributes, and
\item \emph{hybrid-conditioned} generation, combining static attributes with time-varying control variables.
\end{enumerate*}
The unconditional baselines include GMM (Gaussian Mixture Model)~\cite{bishop2006pattern}, TimeGAN~\cite{timegan2021}, TimeVAE~\cite{timevae2022}, Diffusion-TS~\cite{diffusionts2024}, TimeVQVAE~\cite{timevqvae2023}, TimeWeaver~\cite{timeweaver2024}, and EnergyDiff~\cite{energydiff2025}.
Among these methods, TimeVQVAE and TimeWeaver also support static conditioning, while TimeWeaver is the only baseline that supports hybrid conditioning.

Because the baselines do not expose identical conditioning interfaces, each method is evaluated under the regimes it supports natively. 
For generation-quality experiments (Table~\ref{tab:expe1_generation_results}), we report the unconditional variant of each baseline. For downstream-utility experiments (Tables~\ref{tab:tstr_forecasting} and~\ref{tab:tstr_appliance_classification}), we use the conditioned variant of TimeWeaver and \textsc{LoaDiff}, since generating synthetic data matching the target population's static and dynamic attributes is the practically relevant setting; the remaining baselines do not support conditional generation and are evaluated in their only available form.

\subsubsection{Evaluation Metrics}
\label{sec:metrics}

The experiments are structured around four dimensions:
\begin{enumerate*}[(i)]
    \item fidelity and diversity of the generated load curves with respect to the real training data,
    \item privacy, assessing evidence of training-record memorization,
    \item utility for downstream learning tasks, and
    \item controllability under alternative  conditioning scenarios.
\end{enumerate*}

\paragraph{Fidelity and diversity}
We quantify distribution matching between real test data and synthetic samples using:
\begin{itemize}[label=\textbullet, leftmargin=*, labelindent=0pt]
    \item a discriminative 1-NN two-sample test (Discriminative$_{\text{1NN}}$)~\cite{timegan2021}, where values close to $0.5$ indicate that real and generated samples are difficult to distinguish;
    \item embedding-based distances, including a Fréchet distance inspired by the standard FID~\cite{jeha2021psagan}, with ROCKET~\cite{dempster2020rocket} used as the feature extractor in place of an Inception network, and an autocorrelation discrepancy (ACD)~\cite{ni2022sigwasserstein} that captures mismatches in temporal dependence; and
    \item t-SNE projections~\cite{maaten2008visualizing}, used as a qualitative diagnostic of the overlap between real and synthetic distributions.
\end{itemize}
Lower FID and ACD values indicate higher fidelity.

\paragraph{Privacy}
We use distance- and neighborhood-based metrics to assess whether the generator reproduces or memorizes training samples. Nearest-Neighbor Distance Ratio (NNDR) measures the relative distance between a generated sample and its closest real counterparts. 
NeighborsPrivacy relies on a $k$-nearest-neighbor voting scheme, with a reference value of $0.5$ corresponding to a balanced real/synthetic neighborhood structure and, therefore, limited evidence of memorization.



\paragraph{Downstream utility}
We assess whether synthetic data can replace or complement real data for downstream learning. We compare three protocols: Train-on-Real Test-on-Real (TRTR), used as the reference; Train-on-Synthetic Test-on-Real (TSTR), which measures the standalone utility of generated data; and Train-on-Real-and-Synthetic Test-on-Real (TR+STR), which measures their value as a data-augmentation mechanism. 
These protocols are applied to two representative tasks: 
(i) two-day-ahead load forecasting using PatchTST~\cite{nie2023timeseries}; and 
(ii) appliance detection using ROCKET~\cite{dempster2020rocket} and TransApp~\cite{VLDB_TransApp}.



\paragraph{Conditional controllability}
Beyond distribution matching and downstream utility, we test whether \textsc{LoaDiff} responds coherently to controlled changes in its dynamic conditions.
We conduct a counterfactual temperature-sensitivity analysis in Section~\ref{sec:results_temp_sensitivity}, in which winter temperatures are shifted.


\subsubsection{Settings and Computational Resources}

All models use identical 70/15/15\% train/validation/test splits and normalized data, with inverse transformation of the outputs to physical units. 
Experiments run on a DGX-H100 cluster using one NVIDIA H100 GPU (80GB) per training or inference run. 
\textsc{LoaDiff} is trained for up to 300k steps and sampled using standard ancestral DDPM with 100 reverse steps and no respacing, i.e., the full training diffusion schedule. 
Baselines follow their official repositories and original configurations when available. 
For TimeWeaver and EnergyDiff, data are reshaped to $[B,48,365]$ to meet computational constraints while preserving the annual load structure. The GMM uses 10 components with diagonal covariances. 
\newcommand{\pmstd}[2]{#1\,{\tiny$\pm$#2}}

\begin{table}[t]
\centering
\caption{
Results for fidelity, diversity, and privacy across datasets for different baselines. Cells report \emph{mean $\pm$ std} (std to 2 s.f.) over $B=50$ bootstrap resamples of the real/synthetic populations.
NNPrivacy is reported as: $\lvert \mathrm{NNPrivacy}-0.5\rvert$.
Lower is better for all metrics.
Best and second-best results are in \textbf{bold} and \underline{underlined}, respectively.
}
\label{tab:expe1_generation_results}
{\scriptsize
\setlength{\tabcolsep}{3pt}
\renewcommand{\arraystretch}{1.05}

\begin{tabular}{l ccc cc}
\toprule
& \multicolumn{3}{c}{\textbf{Fid. \& Div.}}
& \multicolumn{2}{c}{\textbf{Privacy}} \\
\cmidrule(lr){2-4} \cmidrule(lr){5-6}
Model
& \makecell{Disc$_{1\mathrm{NN}}$\\{\tiny$\to 0.5$}}
& FID $\downarrow$
& ACD $\downarrow$
& NNDR $\downarrow$
& NNPriv. $\downarrow$ \\
\midrule

\multicolumn{6}{l}{\textbf{CER}} \\
GMM
& \textbf{\pmstd{.5000}{.0000}} & \pmstd{.0206}{.0004} & \pmstd{.0101}{.0006} & \pmstd{.0051}{.0002} & \pmstd{.5000}{.0000} \\
TimeVAE
& \pmstd{.6545}{.0172} & \pmstd{.1064}{.0006} & \pmstd{.1594}{.0006} & \pmstd{.0560}{.0003} & \pmstd{.5000}{.0000} \\
TimeGAN
& \pmstd{.8329}{.0261} & \pmstd{.5812}{.0002} & \pmstd{.2111}{.0004} & \pmstd{.0018}{.0002} & \pmstd{.5000}{.0000} \\
TimeVQVAE
& \pmstd{.5130}{.0087} & \pmstd{.0635}{.0012} & \pmstd{.1278}{.0016} & \pmstd{.0329}{.0015} & \pmstd{.5000}{.0000} \\
Diffusion-TS
& \pmstd{.9997}{.0010} & \pmstd{.0408}{.0005} & \pmstd{.0957}{.0006} & \pmstd{.0062}{.0002} & \pmstd{.4991}{.0007} \\
TimeWeaver
& \pmstd{.7182}{.0369} & \pmstd{.0137}{.0008} & \pmstd{.0726}{.0012} & \pmstd{.0086}{.0006} & \textbf{\pmstd{.4845}{.0057}} \\
EnergyDiff
& \underline{\pmstd{.5006}{.0112}} & \underline{\pmstd{.0076}{.0005}} & \underline{\pmstd{.0047}{.0007}} & \underline{\pmstd{.0017}{.0003}} & \pmstd{.5000}{.0000} \\
\textsc{LoaDiff}
& \pmstd{.5104}{.0265} & \textbf{\pmstd{.0073}{.0006}} & \textbf{\pmstd{.0046}{.0010}} & \textbf{\pmstd{.0009}{.0003}} & \underline{\pmstd{.4934}{.0039}} \\

\multicolumn{6}{l}{\textbf{EDF 1}} \\
GMM
& \textbf{\pmstd{.5013}{.0029}} & \pmstd{.0246}{.0006} & \pmstd{.0176}{.0012} & \underline{\pmstd{.0033}{.0007}} & \pmstd{.5000}{.0000} \\
TimeVAE
& \pmstd{.8185}{.0261} & \pmstd{.2130}{.0013} & \pmstd{.1776}{.0019} & \pmstd{.0037}{.0007} & \pmstd{.5000}{.0000} \\
TimeGAN
& \pmstd{.7707}{.0283} & \pmstd{.3344}{.0048} & \pmstd{.2236}{.0035} & \pmstd{.0211}{.0012} & \pmstd{.4962}{.0028} \\
TimeVQVAE
& \underline{\pmstd{.5143}{.0109}} & \pmstd{.0322}{.0025} & \pmstd{.0559}{.0019} & \pmstd{.0220}{.0026} & \pmstd{.4996}{.0010} \\
Diffusion-TS
& \pmstd{1.0000}{.0000} & \pmstd{.0293}{.0012} & \pmstd{.1531}{.0023} & \pmstd{.0140}{.0006} & \pmstd{.4996}{.0010} \\
TimeWeaver
& \pmstd{.8139}{.0376} & \pmstd{.0172}{.0014} & \pmstd{.0747}{.0024} & \pmstd{.0107}{.0008} & \underline{\pmstd{.3787}{.0204}} \\
EnergyDiff
& \pmstd{.5736}{.0392} & \underline{\pmstd{.0103}{.0014}} & \textbf{\pmstd{.0044}{.0012}} & \textbf{\pmstd{.0013}{.0008}} & \pmstd{.4398}{.0236} \\
\textsc{LoaDiff}
& \pmstd{.5481}{.0374} & \textbf{\pmstd{.0096}{.0012}} & \underline{\pmstd{.0093}{.0015}} & \pmstd{.0039}{.0013} & \textbf{\pmstd{.1568}{.0823}} \\

\multicolumn{6}{l}{\textbf{EDF 2}} \\
GMM
& \textbf{\pmstd{.5000}{.0000}} & \pmstd{.0218}{.0015} & \pmstd{.0137}{.0012} & \pmstd{.0051}{.0005} & \pmstd{.4960}{.0011} \\
TimeVAE
& \pmstd{.6249}{.0211} & \pmstd{.1205}{.0017} & \pmstd{.1578}{.0022} & \pmstd{.0427}{.0010} & \pmstd{.4989}{.0012} \\
TimeGAN
& \pmstd{.9952}{.0042} & \pmstd{.3296}{.0080} & \pmstd{.0470}{.0028} & \pmstd{.0030}{.0005} & \pmstd{.4989}{.0012} \\
TimeVQVAE
& \pmstd{.9979}{.0034} & \pmstd{1.3317}{.0132} & \pmstd{.0715}{.0009} & \pmstd{.1891}{.0005} & \pmstd{.5000}{.0000} \\
Diffusion-TS
& \pmstd{1.0000}{.0000} & \pmstd{.0350}{.0013} & \pmstd{.1570}{.0014} & \pmstd{.0104}{.0005} & \pmstd{.5000}{.0000} \\
TimeWeaver
& \underline{\pmstd{.9828}{.0087}} & \pmstd{.0244}{.0017} & \pmstd{.0482}{.0010} & \pmstd{.0100}{.0005} & \underline{\pmstd{.4268}{.0046}} \\
EnergyDiff
& \pmstd{.5127}{.0182} & \textbf{\pmstd{.0121}{.0029}} & \underline{\pmstd{.0120}{.0024}} & \underline{\pmstd{.0028}{.0005}} & \pmstd{.4969}{.0020} \\
\textsc{LoaDiff}
& \pmstd{.6301}{.0262} & \underline{\pmstd{.0123}{.0035}} & \textbf{\pmstd{.0076}{.0011}} & \textbf{\pmstd{.0005}{.0004}} & \textbf{\pmstd{.3295}{.0109}} \\

\bottomrule
\end{tabular}
}
\end{table}

\begin{figure}[tb]
    \centering
    \includegraphics[width=0.8\linewidth]{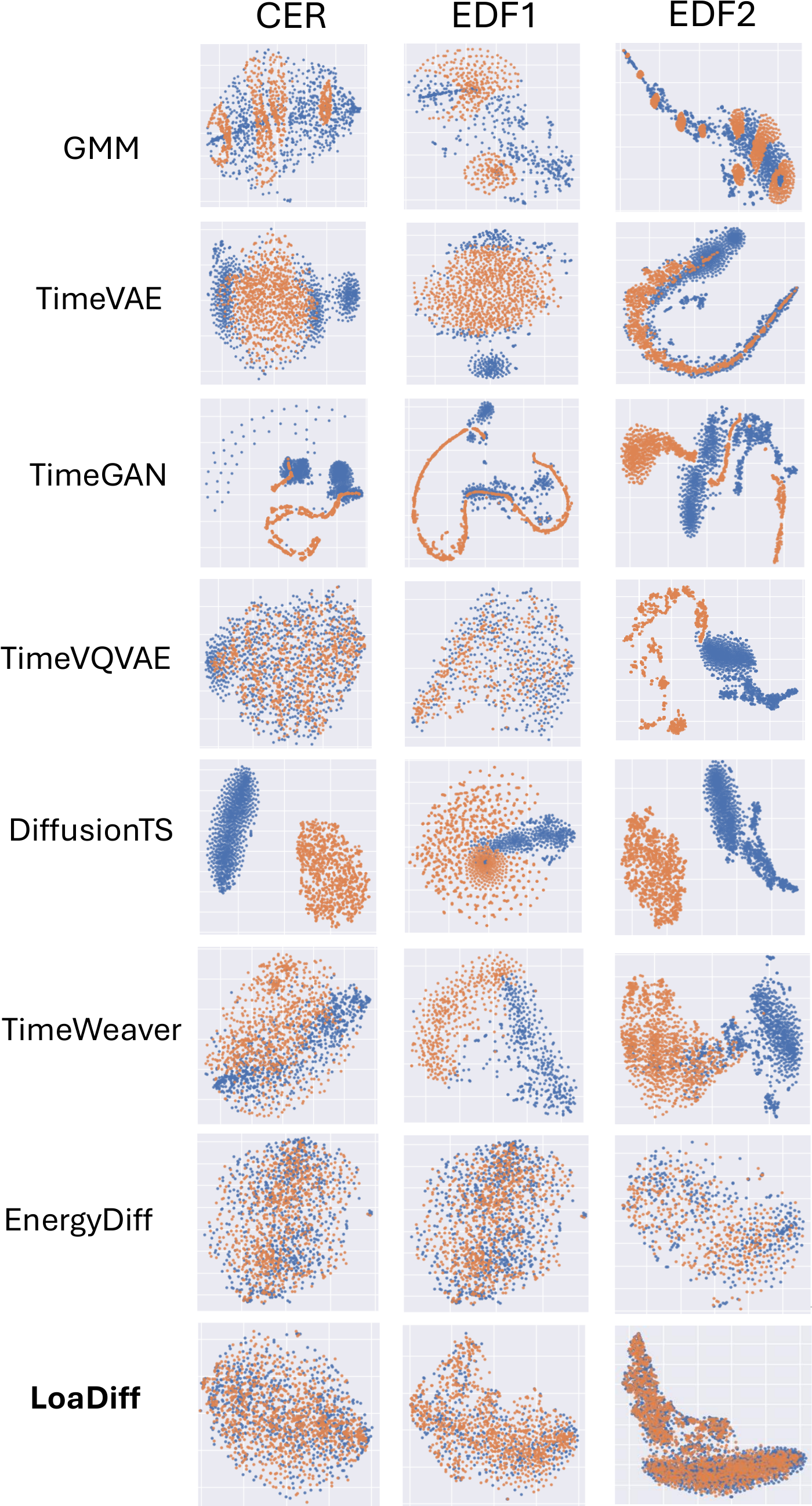}
   \caption{
Two-dimensional t-SNE projections of {\color{blue}real} and {\color{orange}generated} data distributions across different baselines (rows) and datasets (columns). Increased overlap between the two embeddings indicates better coverage of the real data manifold by synthetic samples, reflecting improved fidelity and diversity.
}
    \label{fig:tsne}
    \vspace{-0.4cm}
\end{figure}

\subsection{Results}
\label{sec:results}

We report the results in three stages. 
First, we compare the generated populations in terms of fidelity, diversity, and privacy. 
Second, we evaluate whether synthetic traces preserve the information required by two downstream tasks: load forecasting and appliance detection. 
Third, we examine conditional controllability through counterfactual temperature scenarios. 
Ablations of the main modeling choices are presented in Section~\ref{sec:results_ablations}.

\subsubsection{Data Generation Quality}
\label{sec:results_uncond}

Table~\ref{tab:expe1_generation_results} compares \textsc{LoaDiff} against baselines introduced in Section~\ref{sec:baselines}.

Despite achieving a perfect Discriminative$_{\text{1NN}}$ score, the 10-component GMM only partially recovers the modes of the real data distribution, as observed in the t-SNE visualization in Figure~\ref{fig:tsne}.
This incomplete coverage is consistent with its weaker FID and ACD scores relative to the strongest generators, indicating that important temporal dependencies and higher-order distributional characteristics remain uncaptured.

The strongest diffusion-based variants (\textsc{LoaDiff} and EnergyDiff) jointly occupy the best and second-best FID and ACD scores in every dataset, reflecting closer distributional and temporal alignment with the real data than VAE- and GAN-based methods. This advantage is not shared by every diffusion-based method, however: the classical GMM baseline outperforms Diffusion-TS on both FID and ACD in every dataset, and also outperforms TimeWeaver on ACD in every dataset (and on FID on EDF~2), showing that diffusion-based modeling alone does not guarantee strong fidelity. This ordering does not hold as cleanly for Discriminative$_{\text{1NN}}$ either, where TimeVAE and TimeVQVAE occasionally outperform Diffusion-TS.

\textsc{LoaDiff} achieves the best or second-best FID score in every dataset, and the best or second-best ACD score in every dataset as well. Its Discriminative$_{1\mathrm{NN}}$ score is more variable across datasets: close to $0.5$ on CER ($0.5104$), but ranking mid-table on EDF~1 and EDF~2.
This trend is also supported by the t-SNE visualizations in Figure~\ref{fig:tsne}, where diffusion-based methods exhibit substantially better coverage of the real-data manifold compared to classical generative approaches. Diffusion-TS and TimeWeaver underperform compared to the other diffusion-based methods and, on several metrics, compared to GMM itself, consistent with the Discriminative$_{\text{1NN}}$ exceptions noted above and suggesting challenges in modeling long-range dependencies in very long time series.

The privacy results show no consistent trade-off between generation quality and memorization: higher fidelity does not systematically translate into greater memorization. \textsc{LoaDiff} achieves the lowest NNDR score on CER ($0.0009$) and EDF~2 ($0.0005$); on EDF~1, EnergyDiff instead achieves the lowest NNDR ($0.0013$), with \textsc{LoaDiff} close behind ($0.0039$). On NNPriv, most baselines -- including EnergyDiff on CER and EDF~2 -- sit near the uninformative reference level of $0.5$; \textsc{LoaDiff} moves substantially closer to the ideal deviation of $0$ on EDF~1 and EDF~2, and is edged out only narrowly by TimeWeaver on CER ($0.4934$ vs.\ $0.4845$).

\begin{table}[t]
\centering
\caption{Downstream forecasting utility, reported as \textbf{RMSE / MAE}.
\textsc{TRTR Baseline} uses real data only; \textsc{Synth.} uses synthetic data only; and \textsc{Real + Synth.} combines all synthetic data with the full real training set.
\textbf{Bold} indicates the best synthetic-data generator for each training setting.}
\label{tab:tstr_forecasting}
{\scriptsize
\setlength{\tabcolsep}{3pt}
\renewcommand{\arraystretch}{1.00}
\begin{tabular}{ll cc}
\toprule
\multirow{2}{*}{\textbf{Dataset}}
& \multirow{2}{*}{\textbf{Method}}
& \multicolumn{2}{c}{\textbf{TSTR}} \\
\cmidrule(lr){3-4}
& & \textbf{Synth.} & \textbf{Real + Synth.} \\
\midrule

\multirow{9}{*}{\textbf{CER}}
& TRTR Baseline
& \multicolumn{2}{c}{0.980 / 0.562} \\
\cmidrule(l){2-4}
& GMM
& 1.410 / 0.927
& 0.947 / 0.549 \\
& TimeVAE
& 1.147 / 0.746
& 0.944 / 0.548 \\
& TimeGAN
& 2.608 / 2.072
& 0.946 / 0.548 \\
& TimeVQVAE
& 1.674 / 1.212
& 0.943 / 0.547 \\
& Diffusion-TS
& 1.111 / 0.722
& 0.944 / 0.550 \\
& TimeWeaver
& 1.410 / 0.927
& 0.944 / 0.545 \\
& EnergyDiff
& 1.042 / 0.570
& 0.943 / 0.544 \\
& \textsc{LoaDiff}
& \textbf{0.954 / 0.550}
& \textbf{0.941 / 0.543} \\
\midrule

\multirow{9}{*}{\textbf{EDF 1}}
& TRTR Baseline
& \multicolumn{2}{c}{0.992 / 0.573} \\
\cmidrule(l){2-4}
& GMM
& 1.051 / 0.672
& 0.931 / 0.557 \\
& TimeVAE
& 1.144 / 0.786
& 0.935 / 0.560 \\
& TimeGAN
& 4.436 / 2.649
& 0.950 / 0.560 \\
& TimeVQVAE
& 1.184 / 0.785
& 0.929 / 0.551 \\
& Diffusion-TS
& 1.126 / 0.725
& 0.937 / 0.558 \\
& TimeWeaver
& 1.401 / 0.900
& 0.951 / 0.573 \\
& EnergyDiff
& 0.950 / 0.557
& 0.930 / 0.549 \\
& \textsc{LoaDiff}
& \textbf{0.942 / 0.554}
& \textbf{0.926 / 0.545} \\
\midrule

\multirow{9}{*}{\textbf{EDF 2}}
& TRTR Baseline
& \multicolumn{2}{c}{0.931 / 0.546} \\
\cmidrule(l){2-4}
& GMM
& 1.001 / 0.655
& 0.896 / 0.536 \\
& TimeVAE
& 1.764 / 1.314
& 0.894 / 0.534 \\
& TimeGAN
& 1.324 / 0.980
& 0.921 / 0.579 \\
& TimeVQVAE
& 1.669 / 1.220
& 0.896 / 0.535 \\
& Diffusion-TS
& 1.099 / 0.719
& 0.899 / 0.537 \\
& TimeWeaver
& 1.302 / 0.868
& 0.904 / 0.543 \\
& EnergyDiff
& \textbf{0.915 / 0.536}
& 0.896 / 0.533 \\
& \textsc{LoaDiff}
& 1.027 / 0.577
& \textbf{0.891 / 0.530} \\
\bottomrule
\end{tabular}
}
\end{table}

\subsubsection{Downstream Utility}
\label{sec:results_static}

\definecolor{rocketblue}{HTML}{0072B2}
\definecolor{transapporange}{HTML}{D55E00}
\newcommand{\res}[2]{\textcolor{rocketblue}{#1} / \textcolor{transapporange}{#2}}

\begin{table*}[t]
\centering
\caption{Downstream appliance classification utility results.
Results are reported as balanced accuracy. Each cell reports the results obtained with the two classifiers
\textcolor{rocketblue}{\textbf{ROCKET}} / \textcolor{transapporange}{\textbf{TransApp}}.
The \textsc{TRTR Baseline} uses real training data only.
\textsc{Synth.} uses synthetic data only, while \textsc{Real + Synth.} combines all synthetic data with the full real training set.
\textbf{Bold} indicates the best generator for each classifier, appliance, and training setting; \underline{underlining} indicates the second-best generator.
}
\label{tab:tstr_appliance_classification}
\begin{adjustbox}{width=0.68\textwidth,center}
{
\setlength{\tabcolsep}{2.6pt}
\renewcommand{\arraystretch}{0.96}
\begin{tabular}{l cc cc cc}
\toprule
\textbf{CER} & \multicolumn{2}{c}{\textbf{Cooker}} & \multicolumn{2}{c}{\textbf{Dishwasher}} & \multicolumn{2}{c}{\textbf{Water Heater}} \\
\cmidrule(lr){2-3}\cmidrule(lr){4-5}\cmidrule(lr){6-7}
\textbf{Method} & \textbf{Synth.} & \textbf{Real + Synth.} & \textbf{Synth.} & \textbf{Real + Synth.} & \textbf{Synth.} & \textbf{Real + Synth.} \\
\midrule
TRTR Baseline & \multicolumn{2}{c}{\res{0.589}{0.631}} & \multicolumn{2}{c}{\res{0.663}{0.718}} & \multicolumn{2}{c}{\res{0.582}{0.602}} \\
\cmidrule(l){1-7}
GMM & \res{\underline{0.544}}{\underline{0.555}} & \res{0.574}{0.626} & \res{0.579}{\underline{0.608}} & \res{0.655}{\underline{0.723}} & \res{0.559}{0.480} & \res{\underline{0.586}}{0.605} \\
TimeVAE & \res{0.503}{0.524} & \res{\textbf{0.587}}{0.610} & \res{0.503}{0.509} & \res{0.624}{\textbf{0.725}} & \res{0.500}{0.507} & \res{0.539}{0.600} \\
TimeGAN & \res{0.498}{0.500} & \res{0.580}{0.624} & \res{0.501}{0.500} & \res{0.659}{0.714} & \res{0.500}{0.505} & \res{0.572}{0.586} \\
TimeVQVAE & \res{0.500}{0.497} & \res{0.577}{\underline{0.630}} & \res{0.500}{0.506} & \res{\underline{0.663}}{0.719} & \res{0.500}{0.496} & \res{0.581}{\textbf{0.621}} \\
Diffusion-TS & \res{0.500}{0.521} & \res{0.565}{\textbf{0.630}} & \res{0.500}{0.500} & \res{0.642}{0.672} & \res{0.500}{0.500} & \res{0.580}{0.575} \\
TimeWeaver & \res{0.498}{0.454} & \res{\underline{0.582}}{0.621} & \res{0.500}{0.601} & \res{0.651}{0.718} & \res{0.498}{0.491} & \res{0.585}{\underline{0.613}} \\
EnergyDiff & \res{0.520}{0.541} & \res{0.556}{\textbf{0.630}} & \res{\underline{0.619}}{0.514} & \res{0.628}{0.668} & \res{\underline{0.563}}{\underline{0.556}} & \res{\textbf{0.597}}{0.602} \\
\textsc{LoaDiff} & \res{\textbf{0.570}}{\textbf{0.630}} & \res{0.574}{0.628} & \res{\textbf{0.649}}{\textbf{0.645}} & \res{\textbf{0.665}}{0.704} & \res{\textbf{0.581}}{\textbf{0.564}} & \res{0.583}{0.599} \\

\midrule
\textbf{EDF1} & \multicolumn{2}{c}{\textbf{EV}} & \multicolumn{2}{c}{\textbf{Heater}} & \multicolumn{2}{c}{\textbf{Water Heater}} \\
\cmidrule(lr){2-3}\cmidrule(lr){4-5}\cmidrule(lr){6-7}
\textbf{Method} & \textbf{Synth.} & \textbf{Real + Synth.} & \textbf{Synth.} & \textbf{Real + Synth.} & \textbf{Synth.} & \textbf{Real + Synth.} \\
\midrule
TRTR Baseline & \multicolumn{2}{c}{\res{0.751}{0.771}} & \multicolumn{2}{c}{\res{0.688}{0.706}} & \multicolumn{2}{c}{\res{0.748}{0.784}} \\
\cmidrule(l){1-7}
GMM & \res{0.578}{0.668} & \res{0.632}{0.737} & \res{0.638}{0.649} & \res{0.660}{\underline{0.724}} & \res{0.705}{\underline{0.569}} & \res{0.727}{0.774} \\
TimeVAE & \res{0.516}{0.496} & \res{0.500}{\textbf{0.778}} & \res{0.512}{0.503} & \res{\textbf{0.713}}{0.704} & \res{0.512}{0.500} & \res{0.742}{\underline{0.798}} \\
TimeGAN & \res{0.525}{0.489} & \res{0.533}{\textbf{0.778}} & \res{0.520}{0.560} & \res{0.695}{0.690} & \res{0.512}{0.503} & \res{0.740}{\textbf{0.811}} \\
TimeVQVAE & \res{0.500}{0.395} & \res{0.562}{0.757} & \res{0.500}{0.498} & \res{0.679}{0.708} & \res{0.500}{0.466} & \res{\textbf{0.752}}{0.784} \\
Diffusion-TS & \res{0.500}{0.450} & \res{0.570}{\underline{0.776}} & \res{0.498}{0.497} & \res{0.690}{0.710} & \res{0.500}{0.520} & \res{0.739}{0.785} \\
TimeWeaver & \res{0.327}{\underline{0.706}} & \res{0.750}{0.760} & \res{0.336}{0.490} & \res{0.690}{0.719} & \res{0.283}{0.469} & \res{0.745}{0.781} \\
EnergyDiff & \res{\textbf{0.776}}{0.509} & \res{\underline{0.760}}{0.716} & \res{\textbf{0.683}}{\underline{0.691}} & \res{0.680}{0.675} & \res{\underline{0.711}}{0.550} & \res{0.737}{0.785} \\
\textsc{LoaDiff} & \res{\underline{0.707}}{\textbf{0.758}} & \res{\textbf{0.763}}{0.743} & \res{\underline{0.677}}{\textbf{0.719}} & \res{\underline{0.702}}{\textbf{0.730}} & \res{\textbf{0.731}}{\textbf{0.736}} & \res{\underline{0.746}}{0.788} \\

\midrule
\textbf{EDF2} & \multicolumn{2}{c}{\textbf{Heater}} & \multicolumn{2}{c}{\textbf{AC}} & \multicolumn{2}{c}{\textbf{Water Heater}} \\
\cmidrule(lr){2-3}\cmidrule(lr){4-5}\cmidrule(lr){6-7}
\cmidrule(lr){2-3}\cmidrule(lr){4-5}\cmidrule(lr){6-7}
\textbf{Method} & \textbf{Synth.} & \textbf{Real + Synth.} & \textbf{Synth.} & \textbf{Real + Synth.} & \textbf{Synth.} & \textbf{Real + Synth.} \\
\midrule
TRTR Baseline & \multicolumn{2}{c}{\res{0.899}{0.970}} & \multicolumn{2}{c}{\res{0.645}{0.897}} & \multicolumn{2}{c}{\res{0.890}{0.983}} \\
\cmidrule(l){1-7}
GMM & \res{0.816}{0.794} & \res{0.898}{0.971} & \res{0.507}{0.655} & \res{0.637}{0.903} & \res{0.170}{0.268} & \res{0.884}{0.985} \\
TimeVAE & \res{0.527}{0.538} & \res{0.897}{\underline{0.973}} & \res{0.501}{0.488} & \res{0.637}{0.898} & \res{0.504}{0.500} & \res{0.894}{0.984} \\
TimeGAN & \res{0.504}{0.566} & \res{0.899}{0.970} & \res{0.437}{0.500} & \res{0.636}{0.895} & \res{0.667}{\underline{0.539}} & \res{0.888}{\textbf{0.986}} \\
TimeVQVAE & \res{0.500}{0.500} & \res{0.898}{0.973} & \res{0.500}{\underline{0.678}} & \res{0.638}{0.905} & \res{0.500}{0.500} & \res{\underline{0.895}}{\textbf{0.986}} \\
Diffusion-TS & \res{0.637}{0.778} & \res{0.900}{0.969} & \res{0.502}{0.538} & \res{0.619}{\textbf{0.909}} & \res{\underline{0.756}}{0.500} & \res{0.886}{0.982} \\
TimeWeaver & \res{0.500}{0.663} & \res{0.898}{0.972} & \res{0.500}{0.591} & \res{0.636}{0.895} & \res{0.500}{0.502} & \res{0.888}{0.985} \\
EnergyDiff & \res{\underline{0.864}}{\underline{0.897}} & \res{\underline{0.901}}{0.968} & \res{\underline{0.550}}{0.508} & \res{\underline{0.648}}{0.901} & \res{0.193}{0.487} & \res{0.890}{\underline{0.986}} \\
\textsc{LoaDiff} & \res{\textbf{0.907}}{\textbf{0.942}} & \res{\textbf{0.912}}{\textbf{0.976}} & \res{\textbf{0.621}}{\textbf{0.874}} & \res{\textbf{0.654}}{\underline{0.905}} & \res{\textbf{0.846}}{\textbf{0.864}} & \res{\textbf{0.901}}{0.984} \\

\midrule
\multicolumn{7}{c}{
\makebox[0pt][c]{
\resizebox{0.7\textwidth}{!}{
\begin{tabular}{@{}l c c c c c c c c@{}}
\textbf{}
& \textbf{\textsc{LoaDiff}}
& \textbf{EnergyDiff}
& \textbf{GMM}
& \textbf{TimeVAE}
& \textbf{TimeVQVAE}
& \textbf{TimeGAN}
& \textbf{TimeWeaver}
& \textbf{Diffusion-TS} \\
\cmidrule(lr){2-9}
\textbf{Avg. score}
& \res{\textbf{0.710}}{\textbf{0.766}}
& \res{\underline{0.660}}{0.677}
& \res{0.631}{\underline{0.683}}
& \res{0.595}{0.646}
& \res{0.597}{0.645}
& \res{0.604}{0.651}
& \res{0.576}{0.668}
& \res{0.616}{0.656} \\
\textbf{Avg. rank}
& \res{\textbf{1.67}}{\textbf{2.44}}
& \res{\underline{3.39}}{4.61}
& \res{4.56}{\underline{4.11}}
& \res{4.92}{4.81}
& \res{5.14}{4.81}
& \res{5.11}{5.06}
& \res{5.64}{5.00}
& \res{5.58}{5.17} \\
\end{tabular}
}
}
} \\

\bottomrule
\end{tabular}
}
\end{adjustbox}
\end{table*}

We now evaluate the practical utility of the generated data under the TRTR, TSTR, and TR+STR protocols defined in Section~\ref{sec:metrics}. The first task probes whether synthetic curves preserve predictive temporal structure, while the second tests whether they retain appliance-specific consumption signatures.

\paragraph{Forecasting}
Forecasting evaluates whether synthetic traces preserve temporal dependencies useful for predictive modeling: given a context window $\mathbf{x}_{1:t}$, a model $f_\theta$ predicts the future horizon $\hat{\mathbf{x}}_{t+1:t+H} = f_\theta(\mathbf{x}_{1:t})$, trained by minimizing the mean squared error.

Forecasting is performed over a two-day horizon ($H = 96$) using a two-week context ($C = 512$). 
PatchTST~\cite{nie2023timeseries} is used for all experiments, and performance is measured using RMSE and MAE on instance-normalized data. 
Training uses balanced subsets of real and synthetic data, each containing 1,024 yearly load curves. 
Samples are extracted through overlapping sliding windows, each containing a context of length $C$ and a forecasting horizon of length $H$, with a stride of 96 time steps. For each setting, the model is trained for 500 epochs with a batch size of 1024 and early stopping.
Evaluation is conducted on a held-out test set of 400 real clients. For TimeWeaver and \textsc{LoaDiff}, conditioned samples are generated using the metadata 
of the target test population.

Table~\ref{tab:tstr_forecasting} reports the TRTR, TSTR, and TR+STR scores on CER, EDF1, and EDF2. Models trained on samples generated by EnergyDiff and \textsc{LoaDiff}, which also obtain the strongest generation-quality results in Section~\ref{sec:results_uncond}, achieve forecasting performance comparable to or better than the TRTR reference. 
Across most datasets and evaluation protocols, \textsc{LoaDiff} outperforms the competing baselines. The best overall performance is obtained in the TR+STR setting, indicating that \textsc{LoaDiff} samples provide complementary information and can be used effectively for data augmentation.

\paragraph{Appliance detection}

We assess whether synthetic load curves preserve appliance-specific consumption patterns through a binary appliance-detection task.
For each dataset and target appliance reported in Table~\ref{tab:tstr_appliance_classification}, we generate 2,048 synthetic yearly load curves per method, with an equal number of positive and negative samples.
The same post-processing pipeline is applied to all generated data.
We evaluate two complementary classifiers: ROCKET~\cite{dempster2020rocket}, a lightweight time-series classification baseline, and TransApp~\cite{VLDB_TransApp}, a deep-learning framework specifically designed for appliance detection from smart-meter data.

We apply the TSTR (\textsc{Synth.}) and TR+STR (\textsc{Real + Synth.}) protocols defined in Section~\ref{sec:metrics}.
In both cases, the real validation split is used for model selection and early stopping, and performance is evaluated on the real test split using balanced accuracy to account for class imbalance in the target population.

Table~\ref{tab:tstr_appliance_classification} shows that \textsc{LoaDiff} provides the strongest overall downstream utility.
Aggregated across datasets, appliances, and training settings, \textsc{LoaDiff} ranks first with both classifiers, achieving the highest average balanced accuracy of $0.710 / 0.766$ and the best average rank of $1.67 / 2.44$ with ROCKET / TransApp, respectively.
The corresponding average scores of the strongest competing methods are $0.660 / 0.677$ for EnergyDiff and $0.631 / 0.683$ for GMM.

The distinction is particularly clear in the synthetic-only TSTR setting, where \textsc{LoaDiff} obtains the best result among generators in 16 of the 18 classifier-appliance combinations.
On EDF2, classifiers trained exclusively on \textsc{LoaDiff} samples reach balanced accuracies of $0.907 / 0.942$ for electric heating, $0.621 / 0.874$ for air conditioning, and $0.846 / 0.864$ for water heating.
By contrast, several competing generators yield near-chance performance for at least some appliance labels.
While synthetic data do not universally match the utility of real training samples, these results indicate that \textsc{LoaDiff} is a credible substitute in most settings when access to sensitive household-level data is restricted.

Synthetic curves can also be used to complement real observations.
When \textsc{LoaDiff} samples are added to the real training set, performance improves over the real-only \textsc{TRTR Baseline} in 12 of the 18 classifier--appliance combinations, including all six EDF2 cases.
The improvements are not systematic across every dataset and classifier, but the aggregate results show that \textsc{LoaDiff} is a reliable augmentation strategy rather than solely a synthetic-data replacement mechanism.

\begin{figure*}[tb]
    \centering
    \includegraphics[width=0.89\linewidth]{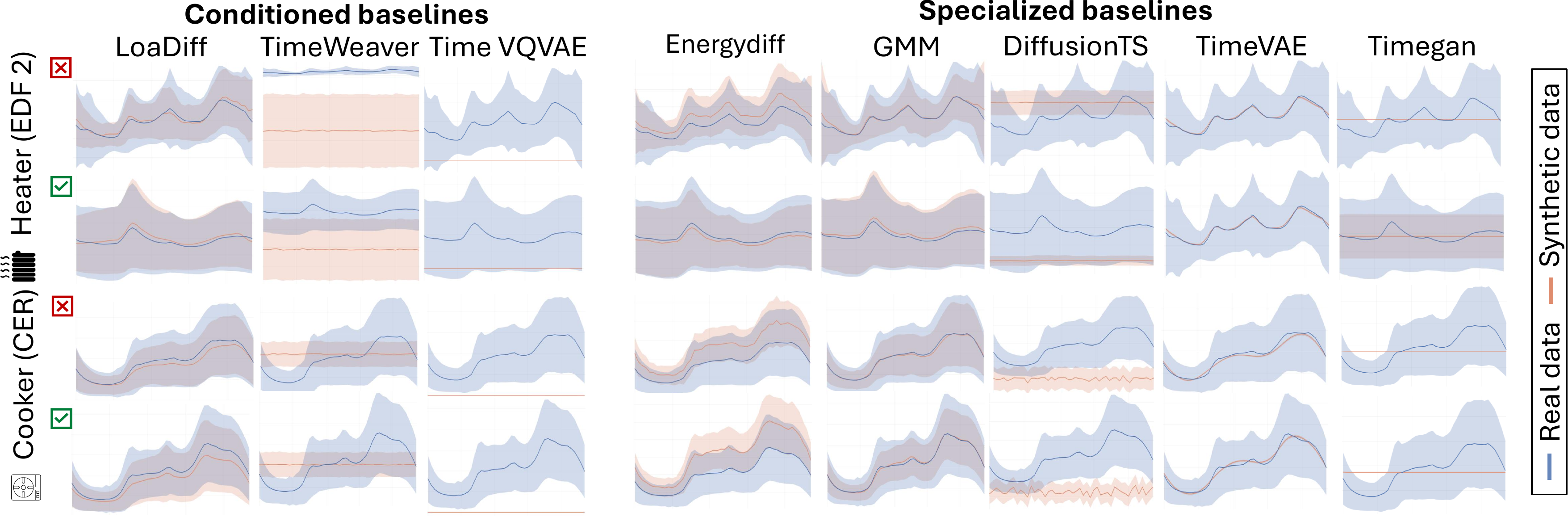}
    \caption{Average daily consumption profiles generated for households with a given appliance label across the evaluated baselines. 
    We compare conditional models with appliance-specific models trained only on load curves associated with the target label.} 
    \label{fig:appliance_profiles}
    \vspace{-0.5cm}
\end{figure*}

Figure~\ref{fig:appliance_profiles} qualitatively compares real and synthetic average daily profiles for households with and without each target appliance.
For electric heating on EDF2 and cookers on CER, \textsc{LoaDiff} most closely reproduces both the overall profiles and the appliance-specific differences between classes, suggesting that its conditioning mechanism preserves meaningful appliance signatures.

\begin{figure}[tb]
\centering
\includegraphics[width=\linewidth]{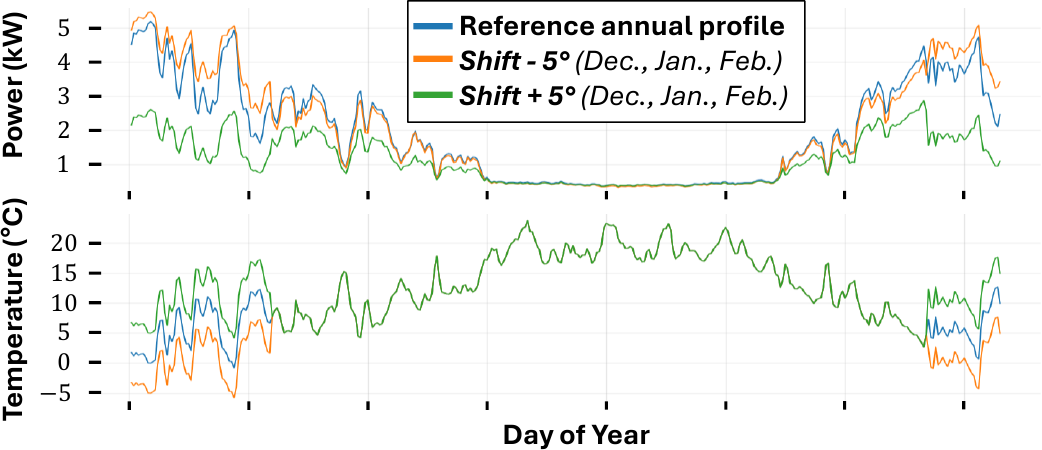}
\caption{Impact of counterfactual winter-temperature shifts on generated yearly load profiles. Top: generated load profiles. Bottom: corresponding temperature profiles.}
\label{fig:temperature_impact}
\vspace{-0.3cm}
\end{figure}

\subsubsection{Conditional Controllability: Temperature Sensitivity}
\label{sec:results_temp_sensitivity}

We assess the sensitivity of \textsc{LoaDiff} to temperature conditioning through a controlled counterfactual experiment. 
To isolate the effect of temperature, we generate yearly profiles with the electric-heating label enabled for all samples. 
We compare the original temperature profile with two scenarios in which winter temperatures are shifted by $-5\,^{\circ}\mathrm{C}$ and $+5\,^{\circ}\mathrm{C}$.
Under the reference profile, the mean winter temperature is $5.8\,^{\circ}\mathrm{C}$ and the mean winter load is $3{,}395.4\,\mathrm{W}$. 
A $-5\,^{\circ}\mathrm{C}$ shift increases the load to $3{,}754.4\,\mathrm{W}$ ($+10.6\%$), whereas a $+5\,^{\circ}\mathrm{C}$ shift reduces it to $1{,}727.7\,\mathrm{W}$ ($-49.1\%$). 
Figure~\ref{fig:temperature_impact} shows the resulting profiles. 
The asymmetric response indicates that \textsc{LoaDiff} is consistent with a nonlinear relationship between temperature and electric-heating demand, though characterizing its precise shape would require evaluating additional intermediate temperature shifts beyond the two directions tested here. 






\subsection{Ablations}
\label{sec:results_ablations}








\begin{table}[t]
\centering
\caption{
Ablation study of \textsc{LoaDiff} on EDF2 dataset.
For Disc$_{1\mathrm{NN}}$, values closer to $0.5$ are better.
FID and ACD are minimized.
Best results are in \textbf{bold}; second-best are \underline{underlined}.
Unless noted otherwise, values are mean $\pm$ std over $n{=}50$ evaluation runs.
}
\label{tab:ablations}
{\scriptsize
\setlength{\tabcolsep}{3.5pt}
\renewcommand{\arraystretch}{1.00}

\begin{threeparttable}
\begin{tabular}{@{} l l c c c @{}}
\toprule
Ablation family
& Configuration
& \makecell{\shortstack{Disc$_{1\mathrm{NN}}$ \\[-1pt] {\tiny $\to 0.5$}}}
& FID $\downarrow$
& ACD $\downarrow$ \\
\midrule

--
& \textbf{Reference}
& \underline{.5175$\pm$.0252}
& \textbf{.0128$\pm$.0039}
& \textbf{.0086$\pm$.0018} \\

\midrule
\multirow[c]{3}{*}{\makecell[l]{Conditioning\\variables}}
& \texttt{no\_static\_cond}
& .5364$\pm$.0253
& \underline{.0129$\pm$.0039}
& .0118$\pm$.0031 \\
& \texttt{no\_temp}
& .5318$\pm$.0226
& .0133$\pm$.0042
& .0129$\pm$.0032 \\
& \texttt{no\_calendar}
& .5188$\pm$.0179
& .0130$\pm$.0039
& .0123$\pm$.0025 \\

\midrule
\multirow[c]{2}{*}{\makecell[l]{Conditioning\\mechanism}}
& \texttt{concat\_cond}
& .5394$\pm$.0165
& .0139$\pm$.0029
& \underline{.0107$\pm$.0025} \\
& \texttt{no\_cfg}
& .5306$\pm$.0167
& .0134$\pm$.0038
& .0127$\pm$.0022 \\

\midrule
\multirow[c]{3}{*}{\makecell[l]{Temporal\\tokenization}}
& \texttt{patch\_24steps}
& .5238$\pm$.0138
& .0136$\pm$.0032
& .0110$\pm$.0022 \\
& \texttt{patch\_336steps}
& \textbf{.5000$\pm$.0000}
& .0349$\pm$.0007
& .0556$\pm$.0020 \\

\bottomrule
\end{tabular}
\end{threeparttable}
}
\end{table}

Table~\ref{tab:ablations} reports single-component ablations of
\textsc{LoaDiff}. The reference configuration provides the best overall
trade-off across the three metrics, with the lowest FID ($0.0128$), the
lowest ACD ($0.0086$), and a near-optimal Disc$_{1\mathrm{NN}}$ score
($0.5175$).

\paragraph{Conditioning variables.}
Removing static appliance labels (\texttt{no\_static\_cond}) leaves FID
essentially unchanged (second-best overall) but degrades Disc$_{1\mathrm{NN}}$
and ACD. Removing temperature conditioning (\texttt{no\_temp}) degrades the
metric profile most severely among this group, while removing calendar
features (\texttt{no\_calendar}) has a milder but still consistent negative
effect, indicating that both exogenous variables contribute to generation
quality.

\paragraph{Conditioning mechanism.}
Replacing AdaLN with concatenation-based conditioning (\texttt{concat\_cond})
degrades all three metrics relative to the reference, despite still
achieving the second-best ACD overall. Disabling classifier-free guidance
(\texttt{no\_cfg}) degrades all three metrics as well, most substantially
ACD, and worsens Disc$_{1\mathrm{NN}}$.

\paragraph{Temporal tokenization.}
Temporal patch size has the largest impact. The reference uses one patch
per calendar day (\texttt{patch\_size=[1,48]}), matching the load curves'
diurnal periodicity, and Table~\ref{tab:ablations} supports this choice
from both sides. Finer patches (\texttt{patch\_24steps}, half-day) raise
FID and ACD above the reference without any compensating benefit. Coarser
patches (\texttt{patch\_336steps}, weekly) obtain the best
Disc$_{1\mathrm{NN}}$ score ($0.5000$) but FID and ACD increase by
$2.7\times$ and $6.5\times$, respectively — a discriminator unable to
tell real from fake alongside collapsing fidelity is the signature of
mode collapse, not better generation, so Disc$_{1\mathrm{NN}}$ must be
read jointly with curve-level fidelity metrics.

\section{Conclusions and Limitations}
We introduced \textsc{LoaDiff}, a conditional diffusion model for generating realistic smart-meter load curves under privacy constraints. Across three independent datasets, \textsc{LoaDiff} consistently ranks among the top methods on fidelity (best or second-best FID and ACD in every setting) and privacy (lowest NNDR on two of three datasets), and delivers the strongest downstream utility of any method compared, winning 16 of 18 classifier, appliance combinations and matching or exceeding real-data forecasting performance in most cases, making it a practical, privacy-preserving substitute for individual smart-meter data. This strength is not absolute: performance varies across datasets and metrics, and no single method, \textsc{LoaDiff} included, dominates on every one, underscoring why evaluation should rely on multiple complementary metrics rather than any single score.

Two of our three datasets are proprietary EDF data and cannot be released; only CER and our code are public, though this means our results are validated on independent populations rather than a single benchmark. Our privacy assessment relies on distance-based heuristics rather than formal guarantees or attack-based evaluations, and our temperature-sensitivity analysis covers only two counterfactual shifts; both should be read as indicative rather than exhaustive.

\section*{Acknowledgments}
This work was supported by EDF R\&D, the French ANRT program, the EU Horizon projects AI4Europe (101070000), TwinODIS (101160009), ARMADA (101168951), and DataGEMS (101188416), and the Greek Ministry of Education project HARSH ($Y\Pi 3TA-0560901$).

%
%
{\footnotesize
\bibliographystyle{IEEEtran}
\bibliography{bibliography}
}
\end{document}